\pdfoutput=1

\documentclass[conference,a4paper]{IEEEtran}
\usepackage{stmaryrd}
\usepackage{amsfonts}
\usepackage{graphicx,times,amsmath,amssymb,float}
\usepackage{amsthm}
\usepackage[vlined,ruled]{algorithm2e}
\usepackage{caption}
\usepackage{epstopdf}
\usepackage{subcaption}

\theoremstyle{definition}
\newtheorem{definition}{Definition}

\theoremstyle{remark}

\usepackage[vlined,ruled]{algorithm2e}
\usepackage{epstopdf}
\usepackage{pdfpages}

\begin{document}
\title{Kernel Alignment for Unsupervised Transfer Learning}

\author{\IEEEauthorblockN{Ievgen Redko}
\IEEEauthorblockA{Laboratoire Hubert Curien, CNRS (UMR 5516)\\
Universit\'e Jean Monnet\\
F-42000, Saint-Etienne, France\\
Email: name.surname@univ-st-etienne.fr}
\and
\IEEEauthorblockN{Youn\`es Bennani}
\IEEEauthorblockA{Laboratoire d'Informatique de Paris-Nord, CNRS (UMR 7030)\\
Universit\'e Paris 13, Sorbonne Paris Cit\'e\\
F-93430, Villetaneuse, France
Email: surname@lipn.fr}
}

\maketitle

\begin{abstract}
The ability of a human being to extrapolate previously gained knowledge to other domains inspired a new family of methods in machine learning called transfer learning. Transfer learning is often based on the assumption that objects in both target and source domains share some common feature and/or data space. In this paper, we propose a simple and intuitive approach that minimizes iteratively the distance between source and target task distributions by optimizing the kernel target alignment (KTA). We show that this procedure is suitable for transfer learning by relating it to Hilbert-Schmidt Independence Criterion (HSIC) and Quadratic Mutual Information (QMI) maximization. We run our method on benchmark computer vision data sets and show that it can outperform some state-of-art methods.
\end{abstract}


%
\IEEEpeerreviewmaketitle

\section{Introduction}
Most research in machine learning is usually concentrated around the setting where a classifier is trained and tested on data drawn from the same distribution. This scenario has already been well investigated and in some tasks supervised approaches have almost no room for improvement. However, building human-like intelligent systems requires them to be able to generalize the discovered patterns to previously unseen domains. This gives rise to a new learning paradigm called transfer learning. Transfer learning is a new learning framework which uses a set of \textit{source} tasks to influence learning and improve performance of \textit{target} task where the same distribution over the source and target samples is not assumed. This important difference between standard setting of machine learning and transfer learning attracts more and more attention nowadays as exploring it helps to understand better under which assumptions and in what way the knowledge can be generalized in human's brain. Intuitively, it is usually assumed that source and target domains should be aligned by learning a new representation of data that maximizes the mutual dependency between them. On the other hand, maximizing the dependence explicitly may lead to a complete loss of auxiliary knowledge that can be a complement to target task. In this case, it becomes crucial to understand at what point one should stop reducing the discrepancy between distributions to preserve the auxiliary knowledge contained in source task that yet remains aligned with target task.

\subsection{Background and related works}
There are three types of transfer learning: (1) supervised or inductive transfer learning (when labeled samples are available in target domain but there can be no labeled instances in the source one); (2) semi-supervised or transductive transfer learning (labeled samples are available only for the source learning task); (3) unsupervised transfer learning (no labeled data both in source and target learning tasks).

According to the survey given in \cite{panyang}, the number of methods dealing with the first two settings of transfer learning drastically exceeds the number of articles dedicated to the last one. Indeed, to the best of our knowledge there are only a couple of algorithms that were proposed to solve this problem: self-taught clustering (STC) presented in \cite{daiyang}, transfer spectral clustering (TSC) (\cite{jiang}) and \cite{journals/ijon/ZhangZ11}. 

The main assumption of STC is that two tasks share a latent feature space that can be used as a ``bridge" for transfer learning. The authors perform co-clustering on source and target data simultaneously, while the two co-clusters share the same feature set. TSC is quite similar to STC, it works in the setting of spectral clustering where the low-dimensional shared embedding for two tasks is measured using the objective of bipartite graph co-clustering.

Another approach that can be related to unsupervised transfer learning is \cite{journals/ijon/ZhangZ11}. The proposed method, however, is an instance of multi-task clustering rather then self-taught clustering. The optimization procedure presented in this work simultaneously minimizes two terms: first represents the sum of Bregman divergence between clusters and data of each task;  second is a regularization term defined as the Bregman divergence between all pairs of partitions. The motivation for this cost function is two-fold - while the first term seeks a qualitative clustering for each task separately, second term takes into account the relationships between clusters of different tasks.

Little research that has been done in this field of machine learning can be explained by the fact that unsupervised transfer learning is an extreme case of the transfer learning paradigm which, nevertheless, occurs in numerous real-world applications. Thus, unsupervised transfer learning becomes a topic of an ongoing interest for further researches.

\subsection{Our contributions}
In this paper, we propose a new unsupervised transfer learning algorithm based on kernel target alignment maximization with application to computer vision problem. To the best of our knowledge, kernel target alignment has never been applied in this context and thus the proposed method presents a novel contribution. 

The rest of this paper is organized as follows: in section 2 we briefly introduce basic notations and describe the approaches used later, in section 3 we are introducing our unsupervised transfer learning algorithm. We present the theoretical analysis of our approach in section 4. In section 5 the proposed approach will be evaluated. Finally, we will point out some ideas about the future extensions of our method in section 6.

\section{Preliminary knowledge}
In this section we describe some basic notations and techniques that are used later. 
We start by introducing the simplest form of Non-negative matrix factorization proposed by \cite{leeseung} and its variations Convex NMF(C-NMF) \cite{dingli} and Kernel NMF \cite{zhangchen}. Each of these methods can be used in order to obtain a partition of data in an unsupervised manner. 

\subsection{Standard NMF}
Given matrix $X \in \mathbb{R} ^ {m \times n}$, standard NMF seeks the following decomposition:
$$X \simeq UH^T, X \in \mathbb{R} ^ {m \times n}, U \in \mathbb{R} ^ {m \times k},H \in \mathbb{R} ^ {n \times k}, \ X,U,H\geq0,$$
where 
\begin{itemize}
\item {columns of $U$ can be considered as basis vectors;} 
\item {columns of $H$ are considered as cluster assignments for each data object;} 
\item {$k$ is the desired number of clusters.} 
\end{itemize}
\subsection{Convex NMF and Kernel NMF}
To develop C-NMF, we consider the factorization of the following form:
$$X \simeq UH^T = XWH^T, W \in \mathbb{R} ^ {n \times k},H \in \mathbb{R} ^ {n \times k}, W,H\geq0.$$
where the column vectors of $U$ lie within the column space of $X$, i.e., $U=XW.$

The natural generalization of C-NMF is Kernel NMF (K-NMF). To ``kernelize" C-NMF we consider a mapping $\phi$ which maps each vector $x_i$ to a higher dimensional feature space, such that:
$$\phi : X \rightarrow \phi(X) = (\phi(x_1),\phi(x_1),...,\phi(x_n)).$$
We obtain the factorization of the following form:
$$\phi(X) \simeq \phi(X)WH^T, X \in \mathbb{R} ^ {n \times m}, W \in \mathbb{R} ^ {m \times k},H \in \mathbb{R} ^ {m \times k}.$$
Each kernel can be described by its Gram matrix. We call a Gram matrix of a given kernel $k$ some symmetric positive-semi-definite matrix $K$. Subsequently the kernel is an inner-product function defined as $K = \phi(X)^T\phi(X).$ We obtain 
$$\phi(X)^T\phi(X) \simeq \phi(X)^T\phi(X)WH^T.$$
Finally, K-NMF is of the form:
$$K \simeq KWH^T, K \in \mathbb{R} ^ {n \times n}, W \in \mathbb{R} ^ {n \times k},H \in \mathbb{R} ^ {n \times k}.$$
The great advantage of K-NMF is that it can deal with not only attribute-value data but also relational data. 
 
\subsection{Kernel Alignment}

Kernel target alignment (KTA) is a measure of similarity between two Gram matrices, proposed in \cite{crist} and defined as follows:
$$\hat{A}(K_1,K_2) = \frac {\left\langle K_1,K_2 \right\rangle _F} {\sqrt {\left\langle K_1,K_1 \right\rangle_F \left\langle K_2,K_2 \right\rangle_F}}.$$
Frobenius inner product is defined as:
$$\left\langle K_1,K_2 \right\rangle _F = \sum_{i,j=1}^{m} {k_1(x_i,x_j)k_2(x_i,x_j)}$$
where $k_1$ and $k_2$ are two kernels, $K_1$ and $K_2$ are two corresponding Gram matrices.

As we can see, it essentially measures a cosine between two kernel matrices.

\subsection{Clustering evaluation criteria}

There are two classes of clustering evaluation metrics: internal and external clustering evaluation indexes. Speaking about unsupervised clustering, we can only use internal metrics because they are based only on the information intrinsic to the data alone. Among them, the most referenced in literature are the following ones: the Bayesian information criteria, Calinski-Harabasz index, Davies-Bouldin index(DBI), Silhouette index, Dunn index and NIVA index. To estimate the effectiveness of clustering we will use one of the most effective (according to \cite{rendon}) clustering indexes, the Davies-Bouldin index.
 This internal evaluation scheme is calculated as follows:
 \begin{align*}
 DBI = \frac{1}{k} \sum_{i=1}^{k} \max_{j:i \neq j} \left( \frac {d(x_i)+d(x_j)}{d{x_i,x_j}} \right),
 \end{align*}
where $k$ denotes the number of clusters, $i$ and $j$ are cluster labels, $d(x_i)$ and $d(x_j)$ are distances to cluster centroids within clusters $i$ and $j$, $d(x_i,x_j)$ is a measure of separation between clusters $i$ and $j$. 
This index aims to identify sets of clusters that are compact and well separated. Smaller value of DBI indicates a ``better" clustering solution.

\section{Our approach} 
\subsection{Motivation}
In this section we describe our method for unsupervised transfer learning. The central idea that we will use to overcome the difference between weakly-related tasks is mainly inspired by a very popular approach used in neuroscience called Representation Similarity Analysis (RSA) \cite{kr:2008RSA}. This method suggests that a proper comparison between different activity patterns in human's brain can be encoded and further compared using dissimilarity matrices. For a given brain region, the authors interpret the activity pattern associated with each experimental condition as a representation. Then, they obtain a representational dissimilarity matrix by comparing activity patterns with respect to all pairs of observations. This approach allows to relate activity patterns between different modalities of brain-activity measurement (e.g., fMRI and invasive or scalp electrophysiology), and between subjects and species. We follow this approach by replacing the dissimilarity matrices of brain activity patterns of different modalities by kernels defined on source and target task samples. Then, we reduce the distance between two distributions by learning a new representation of data for target task in a Reproducing Kernel Hilbert Space (RKHS). This new representation is further factorized using K-NMF in order to find weights of similarities in the transformed instance space. Finally, we use these weights as a ``bridge" for transfer learning on the target task.
\subsection{Kernel target alignment optimization}  
Let us consider two tasks $\mathcal{T}_S$ and $\mathcal{T}_T$ where the corresponding data samples are given by matrices $X_S = \lbrace x_{s_1},x_{s_2},..., x_{s_n}\rbrace \in \mathcal{R}^m$ and $X_T = \lbrace x_{t_1},x_{t_2},..., x_{t_n} \rbrace \in \mathcal{R}^m$. For the sake of convenience, we will consider data sets $X_S$ and $X_T$ with the same number of instances. This inconvenience can be overcome in two ways: by sub-sampling the bigger data set or by using any kind of a bootstrap to increase the size of the smaller data set. 

We start by calculating the Gram matrices $K_S$ and $K_T$ for both source and target tasks, for example, by using a Gaussian kernel function. Calculating $\hat{A}(K_S, K_T)$ gives us an idea of how correlated the initial kernels are. Small value of $\hat{A}(K_S, K_T)$ means that transfer learning will most likely fail as source and target task distributions are too different. In order to find an intermediate kernel $K_{ST}$ that plays the role of an embedding for both tasks, we will now apply the kernel alignment optimization to the calculated kernels $K_S$,$K_T$ that consists in maximizing the unnormalized kernel alignment over $\alpha_n$:
\begin{align*}
& \max_{\alpha_n} \ \left\langle K_S,K_{ST} \right\rangle _F \\[-1mm] 
K_{ST} =  & \sum_{n=1}^k {\alpha_n K_n(x_{t_i},x_{t_j})}, \forall n, \alpha_n \geq 0.
\end{align*}
Normalization in the cost function is omitted compared to the original definition of kernel alignment in section 2 due to the computational convenience as suggested in \cite{Neumann-et-al-ML}.
Matrix $K_{ST}$ represents a linear combination of kernel matrices $K_n$ (any arbitrary set of kernel functions can be used) calculated based on $X_T$. 
There are several methods which can be used to solve this optimization problem. In our work we use the one that was described in \cite{crist}. The others can be found in \cite{ramona} and in \cite{pothin}.
The proposed optimization problem can be rewritten in the following form:
\begin{align*}
\max -& \boldsymbol{ \alpha}^T(K+\lambda I)\boldsymbol{\alpha} + \textbf{f}^T\boldsymbol{\alpha} \\
s.t. & \alpha_n \geq 0, \ \forall n = 1..k,
\end{align*}
where $K(i,j) = \left\langle K_i,K_j \right\rangle_F $ and $f(i) = \left\langle K_i,K_S \right\rangle_F$. In its current form, the maximization procedure presents a quadratic programming (QP) problem and can be solved using any off-shelf QP solver.
For each kernel $K_{ST}$ obtained in the process of alignment optimization, we look for a set of vectors $W_{ST}$ which arises from the K-NMF of $K_{ST}$:
\begin{align*}
K_{ST} \simeq K_{ST}W_{ST}H_{ST}^T.
\end{align*}

This matrix is of a particular interest as it represents the weights of similarities that lead to a good reconstruction of $K_{ST}$ in a nonlinear RKHS. Due to the alignment optimization procedure, it naturally consists of adapted weights of an embedding between two tasks. The information contained in $W_{ST}$ can be used further with C-NMF for the target task in order to find more efficient basis vectors that are weighted based on a ``good" nonlinear reconstruction of transformed instances. The criteria that we use to evaluate if the obtained reconstruction is ``good" or not is Davies-Bouldin index. We recall that this index shows if the clusters are dense and well-separated.      

More formally, we look for a matrix $W_{ST}^*$ that minimizes the Davies-Bouldin index defined in section 2 with respect to target kernel $K_T$:
$$W_{ST}^* = \arg\min_{W_{ST}} DBI(K_{T}).$$
We call this matrix: the ``bridge matrix".
Given that $K_{ST}$ was calculated as a linear combination of kernels of $X_T$ and was brought closer in sense of alignment to $K_S$, $W_{ST}$ naturally incorporate information about geometrical structure of $X_S$ that can help to find better basis vectors in $X_T$.
\subsection{Transfer process using the "bridge matrix"}
Next step is to perform C-NMF of $X_T$ with the matrix of weights fixed to $W_{ST}^*$. We use C-NMF as it allows us to reinforce the impact of $X_T$ on the partition matrix $H_T$. 
$$X_{T} \simeq X_{T}W_{ST}^*H_{T}^T.$$
We call this factorization : the Bridge Convex NMF (BC-NMF). 

Finally, our approach is summarized in Algorithm 1.

\subsection{Complexity}
At each iteration of our algorithm we perform a K-NMF and that makes our algorithm quite time consuming when the number of instances is large. On the other hand, it does not depend on the number of features that makes its usage attractive for tasks from high-dimensional spaces. The complexity of K-NMF is of order $n^3 + 2m(2n^2k+nk^2) + mnk^2 $ for a Gram matrix $K \in \mathbb{R} ^ {n \times n}$, where $m$ is a number of iterations used for K-NMF to converge (usually, $m \approx 100$), $k$ - is a desired number of clusters. Then, this expressions should be multiplied by $t$ - the number of iterations needed to optimize the alignment between two kernels. Finally, we obtain the following order of complexity: $t(n^3 + 2m(2n^2k+nk^2) + mnk^2)$.

It should be noted that in real-life tasks the quantity of data in source domain is often greater than in the target one. In order to decrease the computational effort of BC-NMF we propose to proceed a data treatment in the parallel fashion. We split data into several parts and obtain an optimal result for each of them. After that, we use any arbitrary consensus approach (for example, Consensus NMF described in \cite{dingjor}) to calculate the final result which is close to all the partitions obtained. 

\vspace{-2mm}
\begin{algorithm}
\small
 \SetKwInOut{Input}{input}\SetKwInOut{Output}{output}
 \Input{$X_S$ - source domain data set, $X_T$ - target domain data set, $r$ - number of clusters, $n_{iter}$  - number of iterations}
 \Output{$H_{ST}*$ - partition matrix, $W_{ST}^*$ - "bridge matrix"}
 Initialize $K_S$, $K_T$\;
 $K_S \leftarrow kernel(X_S,X_S,\sigma)$\;
 $K_T \leftarrow kernel(X_T,X_T,\sigma)$\;
 $\hat{A}_{init} \leftarrow \hat{A}(K_S,K_T)$\; 
 \For{$i\leftarrow 1$ \KwTo $n_{iter}$}{
    $K_{ST} \leftarrow alignment \ optimization(K_S,K_T)$\;
    $W_{ST} \leftarrow  \text{K-NMF}(K_{ST},r)$\;
 }
 $H_{ST}^* \leftarrow \text{CNMF}(X_T,W_{ST}^*,r)$\;
 \caption{\small Bridge Convex NMF (BC-NMF)}
\end{algorithm}
\vspace{-2mm}

\section{Theoretical analysis}
In this section, we present the relationships between KTA and two quantities commonly used in transfer learning and domain adaptation problems, namely: Hilbert Schmidt Independence Criterion (HSIC) \cite{Gretton:2005:MSD:2101372.2101382} and Quadratic Mutual Information.  
\subsection{Hilbert-Schmidt independence criterion}
We start with a definition of a mean map and its empirical estimate. 
\begin{definition}
Let $k:\mathcal{X}\times\mathcal{X} \rightarrow \mathcal{R}$ be a kernel in the RKHS $\mathcal{H}_k$ and $\phi(x) = k(x,\cdot)$. Then, the mapping $\mu[p] = \mathbb{E}_{x\sim p}[\phi(x)]$ is called a mean map. 
Its empirical value is given by the following estimate $\mu[X] = \frac{1}{m}\sum_{i=1}^m\phi(x),$
where $X = \lbrace x_1,...,x_m \rbrace$ is drawn iid. from $p$. 
\end{definition}
If $\mathbb{E}_x[k(x,x)]<\infty$ then $\mu[p]$ is an element of RKHS $\mathcal{H}_k$. According to Moore-Aronszajn theorem, the reproducing property of $\mathcal{H}_k$ allows us to rewrite every function $f \in \mathcal{H}_k$ in the following form: $\langle \mu[p],f \rangle_{\mathcal{H}_k} = \mathbb{E}_x[f(x)]$. 
We now give the definition of HSIC.

\begin{definition}
Let $k(x,x')$ and $l(y,y')$ be bounded kernels with associated feature maps $\phi:\mathcal{X} \rightarrow \mathcal{F}$,  $\psi:\mathcal{Y} \rightarrow \mathcal{G}$ and let $(x,y)$ and $(x',y')$ be independent pairs drawn from the joint distribution $p_{xy}$. Then HSIC is defined as follows: 
\scalebox{0.93}{\parbox{\linewidth}{%
\vspace{-2mm}
\begin{multline*}
$$HSIC(p_{xy},\mathcal{F},\mathcal{G}) = \Vert\mathcal{C}_{xy}\Vert^2 = \mathbb{E}_{x,x',y,y'}\left[k(x,x')l(y,y')\right] + \\ 
+\mathbb{E}_{x,x'}\left[k(x,x')\right] \mathbb{E}_{y,y'}\left[ l(y,y')\right]  +  \mathbb{E}_{x,y}\left[\mathbb{E}_{x'}\left[k(x,x')\right] \mathbb{E}_{y'}\left[l(y,y')\right]\right],$$
\end{multline*}}}
where $\mathcal{C}_{xy} = \mathbb{E}_{x,y}\left[ (k(x,\cdot) - \mu[p])\otimes (k(y,\cdot) - \mu[q]) \right]$ is cross-covariance operator.
\end{definition}
Its biased estimate can be calculated from a finite sample using following equation:
$$\hat{HSIC} = \frac{1}{m^2}tr(KHLH),$$
where $K_{ij} = k(x_i,x_j)$, $L_{ij} = l(y_i,y_j)$ and $H = I - \frac{1}{m}\mathbf{1}\mathbf{1}^T$ is a centering matrix projecting data to a space orthogonal to to the vector $\mathbf{1}$.

From this we can see that KTA coincide with the biased estimate of HSIC when centered kernels are used. It shows that KTA is a suitable choice for transfer learning algorithms as its maximization increases iteratively the dependence between source and target distributions. Furthermore, cross-covariance operator has already proved to be efficient when applied in domain adaptation problem for target and conditional shift correction \cite{icml2013_zhang13d}.
\subsection{Quadratic mutual information}
Another important point is the equivalence between KTA and Information-Theoretic Learning (ITL) estimators \cite{principe}. We define the inner-product between two pdfs as a bivariate function on the set of square intergrable probability density functions:
$$\mathcal{V}(p,q) = \int p(x)q(x)dx.$$ 
It is easy to show that $\mathcal{V}(p,q)$ is symmetric and non-negative definite and thus according to Moore-Aronszajn theorem, there exists a unique RKHS $\mathcal{H}_v$ associated with $\mathcal{V}(p,q)$. We further define Quadratic Mutual Information (QMI):
$$QMI(x,y) = \iint(p(x,y) - p(x)p(y))^2dxdy.$$
In order to establish a connection between KTA and QMI, we can use the equivalence between $\mathcal{H}_v$ and $\mathcal{H}_k$ established in \cite{principe} through Parzen window estimation \cite{parzen1962estimation}. Parzen window estimator of given probability density functions $p(x)$,$p(y)$ and $p(x,y)$ is defined as follows:
\begin{align*}
\hat{p}(x) = \frac{1}{m}\sum_{i=1}^m k_x(x-x_i), \ \hat{p}(y) = \frac{1}{m}\sum_{i=1}^m k_y(y-y_i),\\[-2mm]
\hat{p}(x,y) = \frac{1}{m}\sum_{i=1}^m k_x(x-x_i)k_y(y-y_i).
\end{align*} 
This leads to the following result:
$$\hat{QMI}(x,y) = \Vert \hat{p}(x,y) - \hat{p}(x)\hat{p}(y)\Vert^2 = \frac{1}{m^2}tr(KHLH),$$
where kernel matrices $K$ and $L$ are calculated with respect to Parzen window kernels used for estimation. Once again, we see that KTA with centered kernels is equal to QMI estimation when the Gram matrices $K$ and $L$ are defined as inner-products of Parzen window kernels.

We also note that STC \cite{daiyang} is based on mutual information maximization. The latter was used to perform co-clustering of target and auxiliary data with respect to a shared set of features. Another example where mutual information was used for domain adaptation is \cite{GongGS13}. Thus, we may conclude that the established relationships allow us to assume that KTA can be effective when used for transfer learning.
\section{Experimental results}
In this section we evaluate our approach and analyze its behavior on some popular computer vision data sets. 
\subsection{Baselines and setting}
We choose the following baselines to evaluate the performance of our approach:
\begin{itemize}
\item C-NMF on target data only;
\item K-NMF using each kernel from the set of base kernels used for KTA maximization (``Kernel alone"); 
\item Transfer Spectral Clustering (TSC);
\item Bridge Convex-NMF (BC-NMF).
\end{itemize}
Using C-NMF we can directly factorize matrix $X_T$ as:
$$X_{T} \simeq X_{T}W_{T}H_{T}^T$$
and consider matrix $H_T$ as an initial partition which could be obtained without taking into account the knowledge from the source task. Accuracy obtained on this partition gives us the ``No transfer" value. This particular choice of the baseline can be explained by the fact that our approach is, basically, C-NMF but with a weight matrix $W_{T}$ learned using kernel alignment optimization. Thus, if we are able to increase the accuracy of classification compared to this baseline it will be only due to the efficiency of our approach. 

On the other hand, we also give the maximum value of accuracy achieved for a set of kernels that we use in the optimization of KTA. We chose the following kernel functions: (1) Gaussian kernels with bandwidth varying between $2^{-20}$ to $2^{20}$ with multiplicative step-size of $2$; homogeneous polynomial kernels with the degree varying from $1$ to $3$. We call this ``kernel alone" value as it presents the result of applying K-NMF to a given kernel without taking into account the auxiliary knowledge. Source task kernel was calculated using linear kernel.

Finally, we compare out method to TSC that according to the experimental results presented in \cite{jiang} outperforms both STC and Bregman multitask clustering (BMC). To define the number of nearest neighbors needed to construct the source and target graphs, we perform cross-validation for $k \in [5;100]$ and report the best achieved accuracy value. As suggested in the original paper, we set $\lambda = 3$ and the step length $t = 1$.

We will use accuracy to evaluate the performance of chosen algorithms. It is defined as:
$$Accuracy = \frac{\vert \textbf{x}: \textbf{x} \in \mathcal{D} \wedge \hat{y}(\textbf{x}) = y(v) \vert}{\vert \textbf{x}: \textbf{x} \in \mathcal{D} \vert},$$
where $\mathcal{D}$ is a data set, and $y(\textbf{x})$ is the truth label of $\textbf{x}$ and $\hat{y}(\textbf{x})$ is the predicted label of $\textbf{x}$. 

\subsection{Data sets} 
We evaluate the performance of our approach on the Office \cite{Saenko:2010:AVC:1888089.1888106}/Caltech \cite{Gopalan:2011:DAO} data set which consists of four classification tasks: 
\begin{itemize}
\item Amazon (A) - images from online merchants (958 images with 800 features from 10 classes);
\item Webcam (W) - set of low-quality images by a web camera (295 images with 800 features from 10 classes);
\item DSLR (D) - high-quality images by a digital SLR camera (157 images with 800 features from 10 classes);
\item Caltech (C) - famous data set for object recognition (1123 images with 800 features from 10 classes). 
\end{itemize}

Sample images from keyboard and backpack categories of each domain are presented in Figure 1.  
\begin{figure}[ht]
\vspace{-2mm}
\begin{subfigure}[b]{0.055\textwidth}
  \centering
  \includegraphics[width=\linewidth]{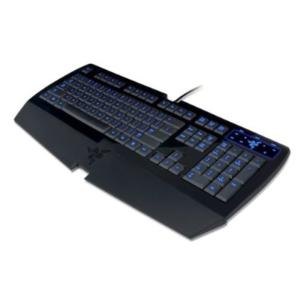}  
  \label{fig:sfig1}
\end{subfigure}%
\begin{subfigure}[b]{0.056\textwidth}
  \centering
  \includegraphics[width=\linewidth]{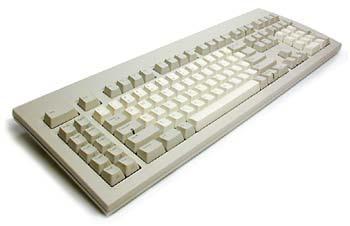}  
  \label{fig:sfig2}
\end{subfigure}
\begin{subfigure}[b]{0.056\textwidth}
  \centering
  \includegraphics[width=\linewidth]{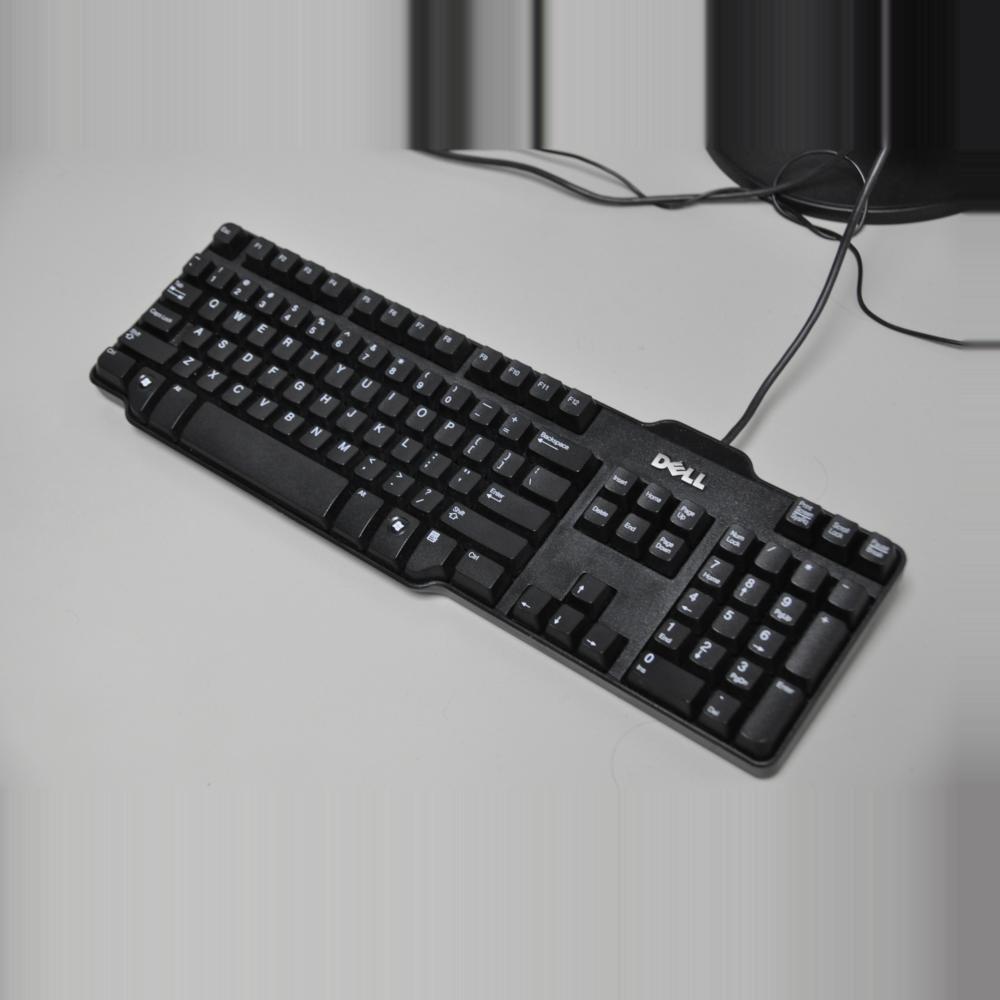}  
  \label{fig:sfig3}
\end{subfigure}%
\begin{subfigure}[b]{0.056\textwidth}
  \centering
  \includegraphics[width=\linewidth]{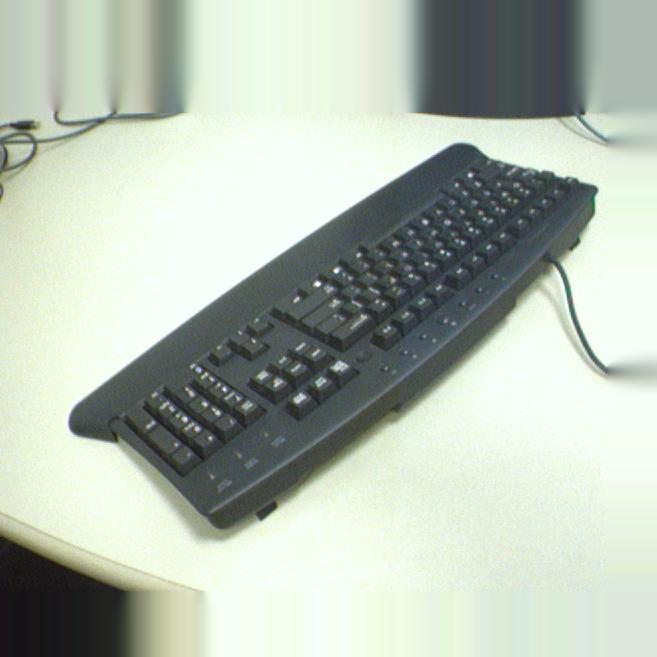}  
  \label{fig:sfig4}
\end{subfigure}
\begin{subfigure}[b]{0.056\textwidth}
  \centering
  \includegraphics[width=\linewidth]{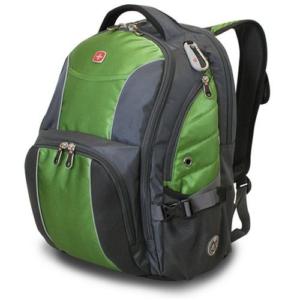}  
  \label{fig:sfig5}
\end{subfigure}%
\begin{subfigure}[b]{0.056\textwidth}
  \centering
  \includegraphics[width=\linewidth]{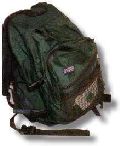}  
  \label{fig:sfig6}
\end{subfigure}
\begin{subfigure}[b]{0.056\textwidth}
  \centering
  \includegraphics[width=\linewidth]{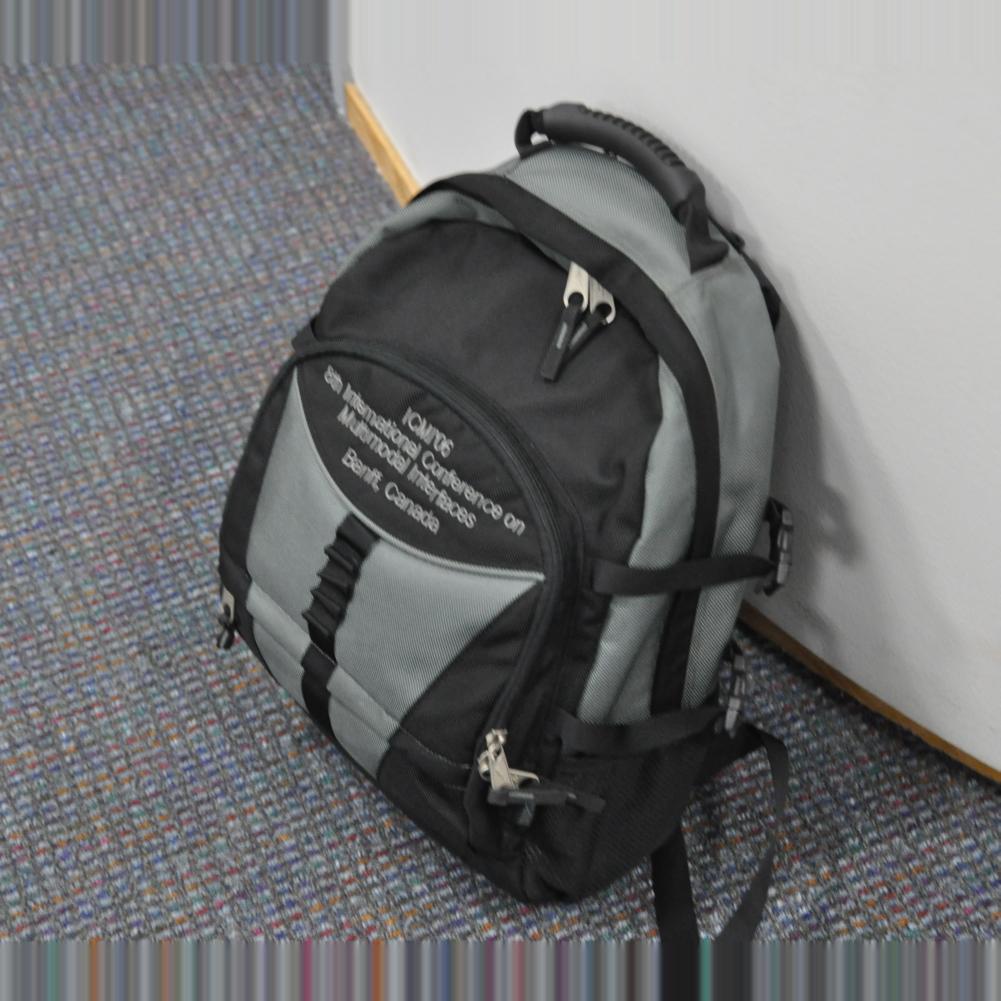}  
  \label{fig:sfig7}
\end{subfigure}%
\begin{subfigure}[b]{0.056\textwidth}
  \centering
  \includegraphics[width=\linewidth]{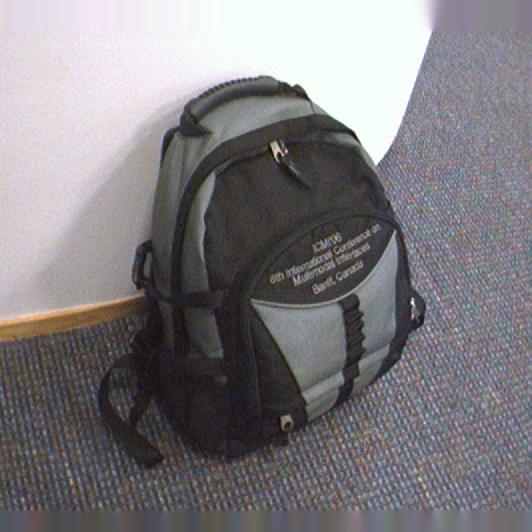}  
  \label{fig:sfig8}
\end{subfigure}
\caption{\small Examples of keyboard and backpack images from Amazon, Caltech, DSLR and Webcam data sets from left to right.}
\vspace{-2mm}
\label{fig:fig}
\end{figure}
This set of domains leads to 12 transfer learning scenarios, e.g., C $\rightarrow$ A, C $\rightarrow$ D, C $\rightarrow$ W, ..., D $\rightarrow$ W. 
\subsection{Results}
In Table 1 we can see the results of the experimental tests of our approach for transfer between two different domains where bold and underlined numbers stand for the best and second best results respectively.
\renewcommand{\arraystretch}{1.1}
\vspace{-2mm}
\begin{table}[ht]
\centering
\caption{Classification accuracy on Office/Caltech data set}
\begin{tabular*}{\columnwidth}{@{\extracolsep{\stretch{1}}}*{5}{c}@{}}
\hline\noalign{\smallskip}
\textbf{Domain pair} & \textbf{C-NMF} & \textbf{Kernel alone} & \textbf{TSC} & \textbf{BC-NMF}\\
\noalign{\smallskip}
\hline
\noalign{\smallskip}
C $\rightarrow$ A & 33.24   & 40.34	& \underline{43.32}	& \textbf{64.88}\\
C $\rightarrow$ W & 46.78	& \underline{56.00}	& 52.54	& \textbf{60.69}\\
C $\rightarrow$ D & 46.5	& 47.33	& \underline{54.14}	& \textbf{81.33}\\
A $\rightarrow$ C & 24.89	& 35.33	& \underline{46.03}	& \textbf{59.29}\\
A $\rightarrow$ W & 46.78	& \underline{56.00}	& 53.22	& \textbf{60.69}\\
A $\rightarrow$ D & 46.5	& 47.33	& \underline{51.59}	& \textbf{76.0}\\
W $\rightarrow$ C & 24.89	& 35.33	& \textbf{62.71}	& \underline{58.97}\\
W $\rightarrow$ A & 33.24	& 40.34	& \underline{61.36}	& \textbf{77.93}\\
W $\rightarrow$ D & 46.5	& 47.33	& \underline{59.66}	& \textbf{76.0}\\
D $\rightarrow$ C & 24.89	& 35.33	& \textbf{54.14}	& \underline{52.0}\\
D $\rightarrow$ A & 33.24	& 40.34	& \underline{54.78}	& \textbf{78.0}\\
D $\rightarrow$ W & 46.78	& \underline{56.00} & 55.59 & \textbf{70.0}\\
\noalign{\smallskip}
\hline
\noalign{\smallskip}
\end{tabular*}
\vspace*{-3mm}
\end{table}       
 
From the results, we can see that our algorithm BC-NMF significantly outperforms TSC in 10 transfer learning scenarios. Furthermore, in some cases TSC achieves lower accuracy values than the ``kernel alone" setting. This can be explained by the fact that the clusters of the corresponding tasks are not well separable in the initial feature space and thus a nonlinear projection of features to a new RKHS can be beneficial. We also note that using a single kernel from the set of base kernels does not lead to good performance when compared to BC-NMF, while the learned combination of base kernels improves the overall classification accuracy considerably. Finally, comparing the obtained results with C-NMF applied to target data only clearly shows that the improved performance is due to the transfer as the only difference between BC-NMF and C-NMF lies in the learned weight matrix $W$.
 
In conclusion, we analyze two cases where TSC achieves better clustering results than BC-NMF. We remark that in these two cases Caltech10 plays the role of the target domain. We further notice that the overall performance of both C-NMF and ``kernel alone" approaches on Caltech10 is rather weak compared to their performance on Amazon, DSLR and Webcam tasks. We recall that both C-NMF and K-NMF assume that the basis vectors lie in the column space of their instance space while it is not necessarily true. However, if the source task data set is large enough, our approach is still able to improve the performance using the auxiliary knowledge (i.e., A $\rightarrow$ C) while when it is not the case (i.e., W $\rightarrow$ C, D $\rightarrow$ C) BC-NMF may need a larger variety of base kernels to learn a good weight matrix $W$ or more instances from the source data set. 

Figure 2 presents the learning curves of BC-NMF on transfer from DSLR and Caltech domains (results for other domains are presented in the Supplementary material). We plotted the red bar to indicate where the optimal weight matrix $W$ was obtained. It can be noticed that the proposed strategy to choose $W_{ST}$ does not always lead to the best possible results but still performs reasonably well.

\begin{figure}[!ht]
\vspace{-3mm}
\begin{subfigure}[b]{0.16\textwidth}
  \centering
  \includegraphics[width=\linewidth]{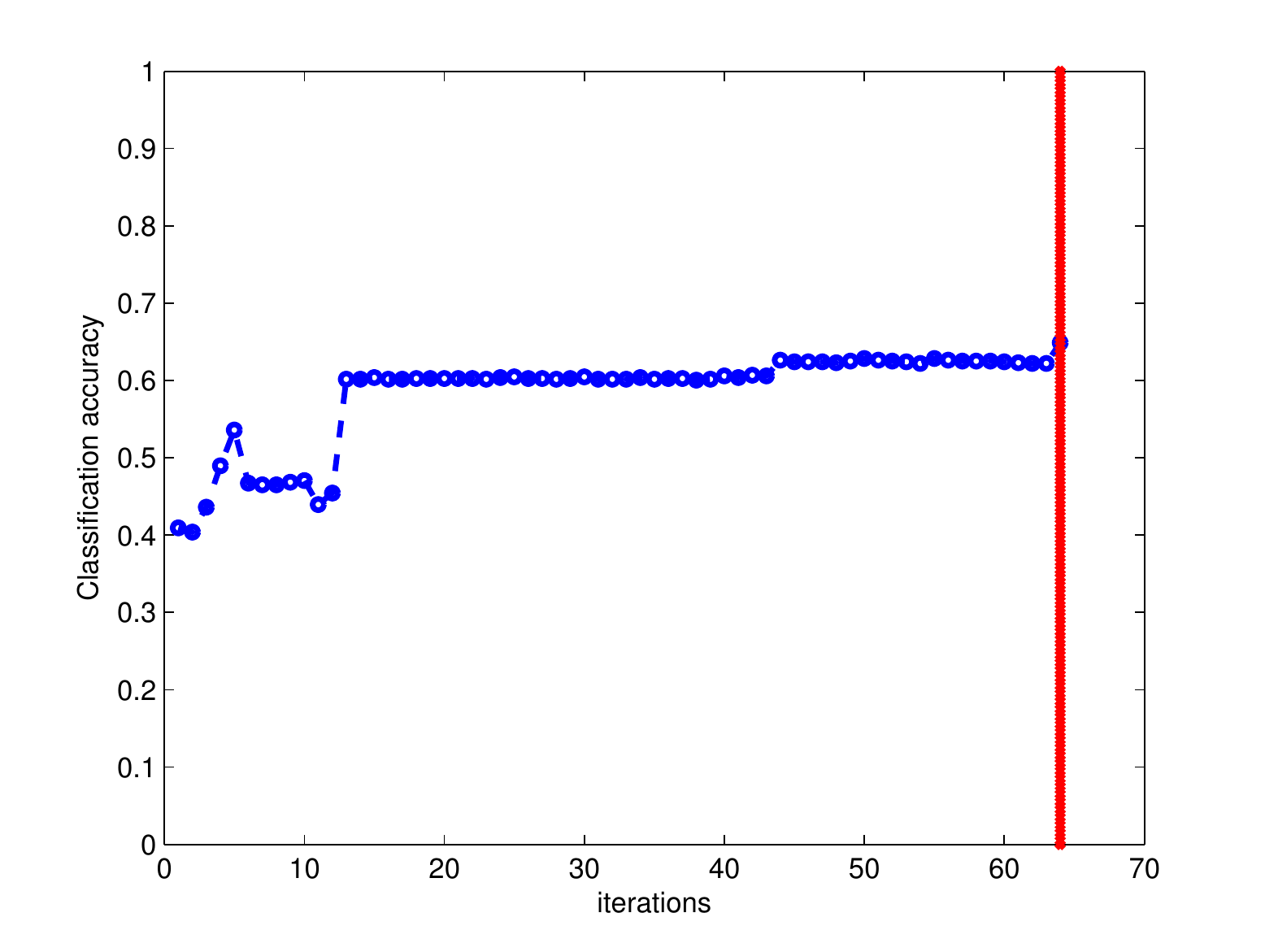}
  \caption{C $\rightarrow$ A}  
  \label{fig:1}
\end{subfigure}%
\begin{subfigure}[b]{0.16\textwidth}
  \centering
  \includegraphics[width=\linewidth]{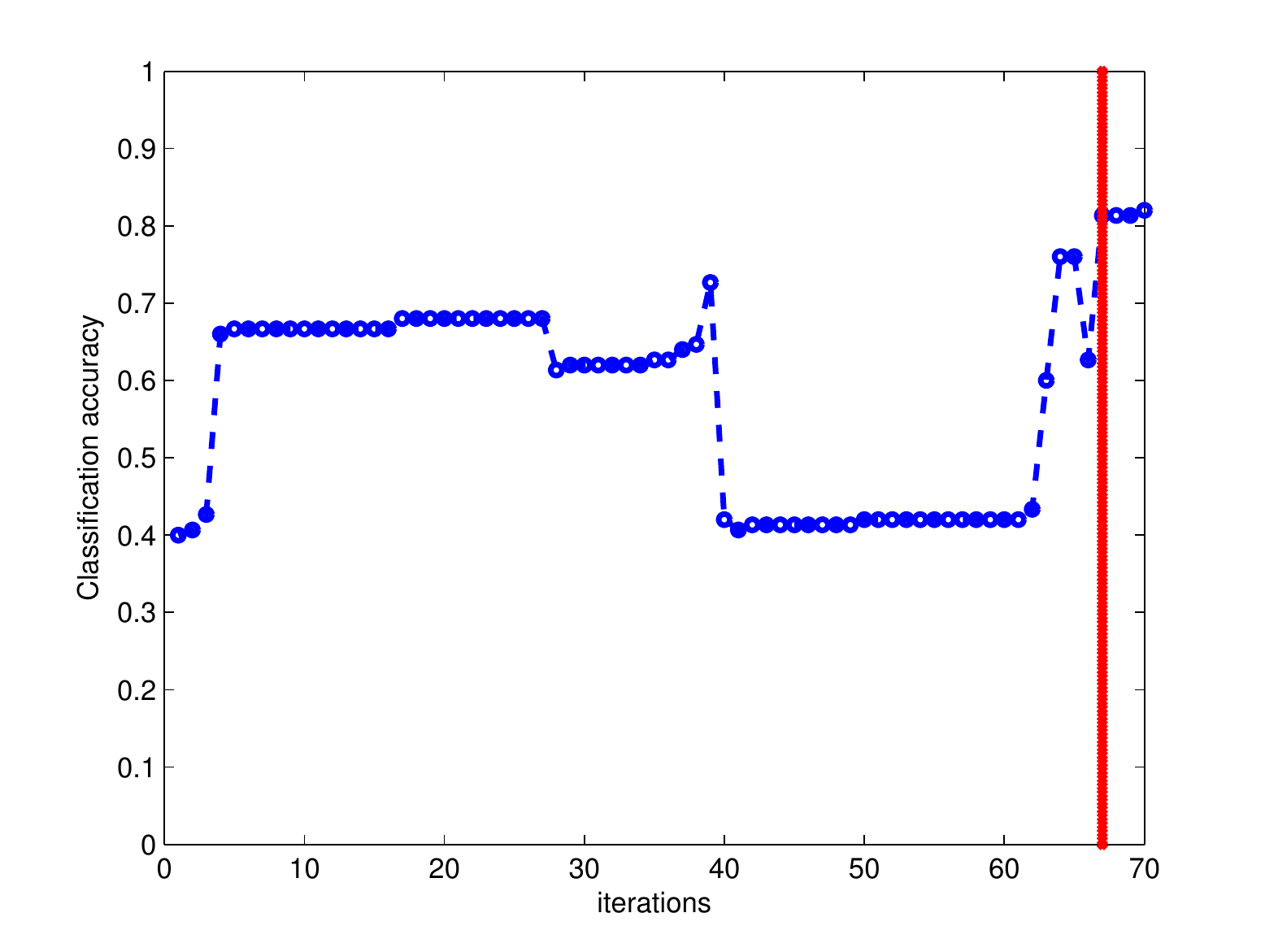} 
  \caption{C $\rightarrow$ D} 
  \label{fig:2}
\end{subfigure}
\begin{subfigure}[b]{0.16\textwidth}
  \centering
  \includegraphics[width=\linewidth]{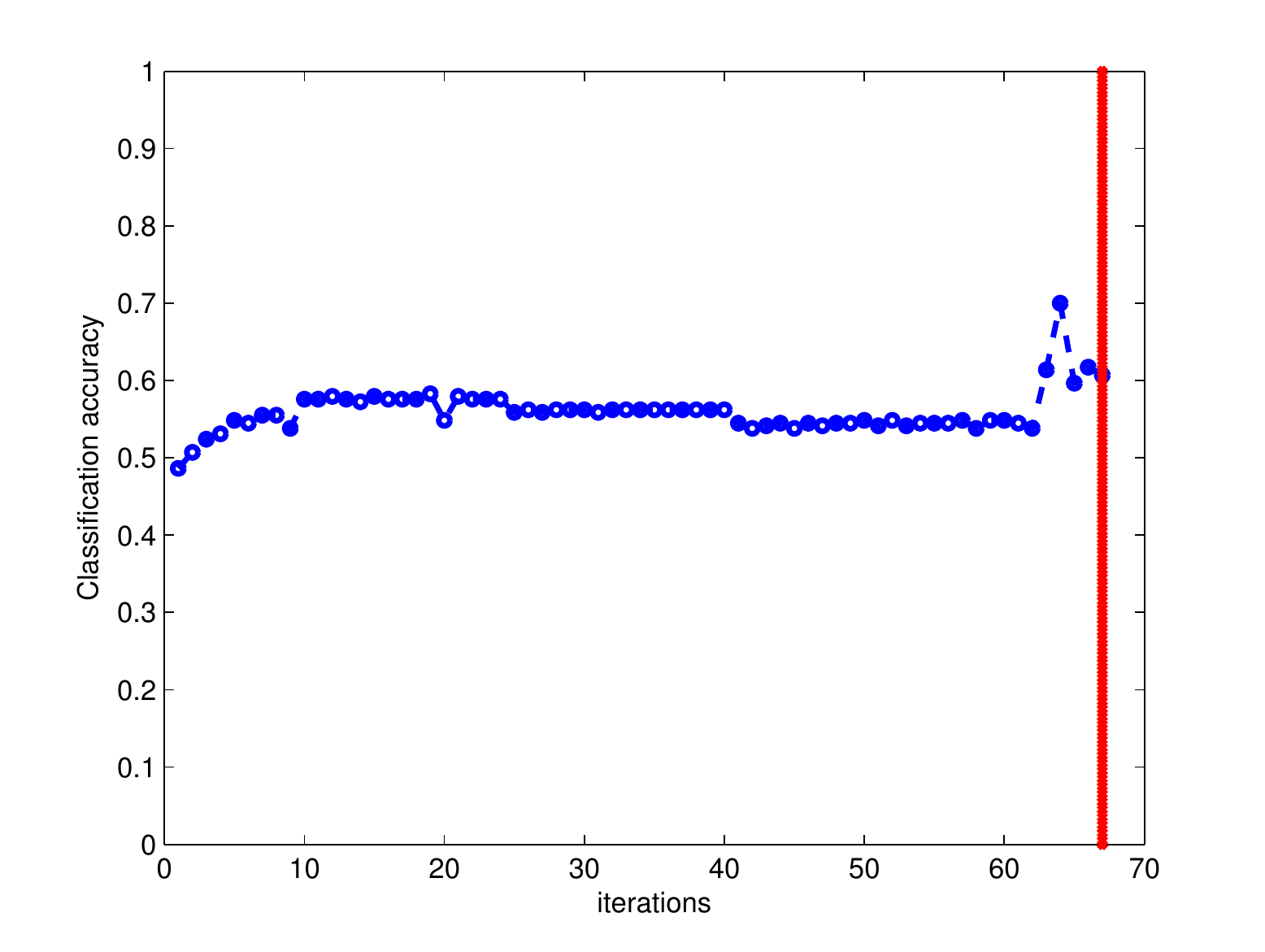}  
  \caption{C $\rightarrow$ W}
  \label{fig:3}
\end{subfigure}
\begin{subfigure}[b]{0.16\textwidth}
  \centering
  \includegraphics[width=\linewidth]{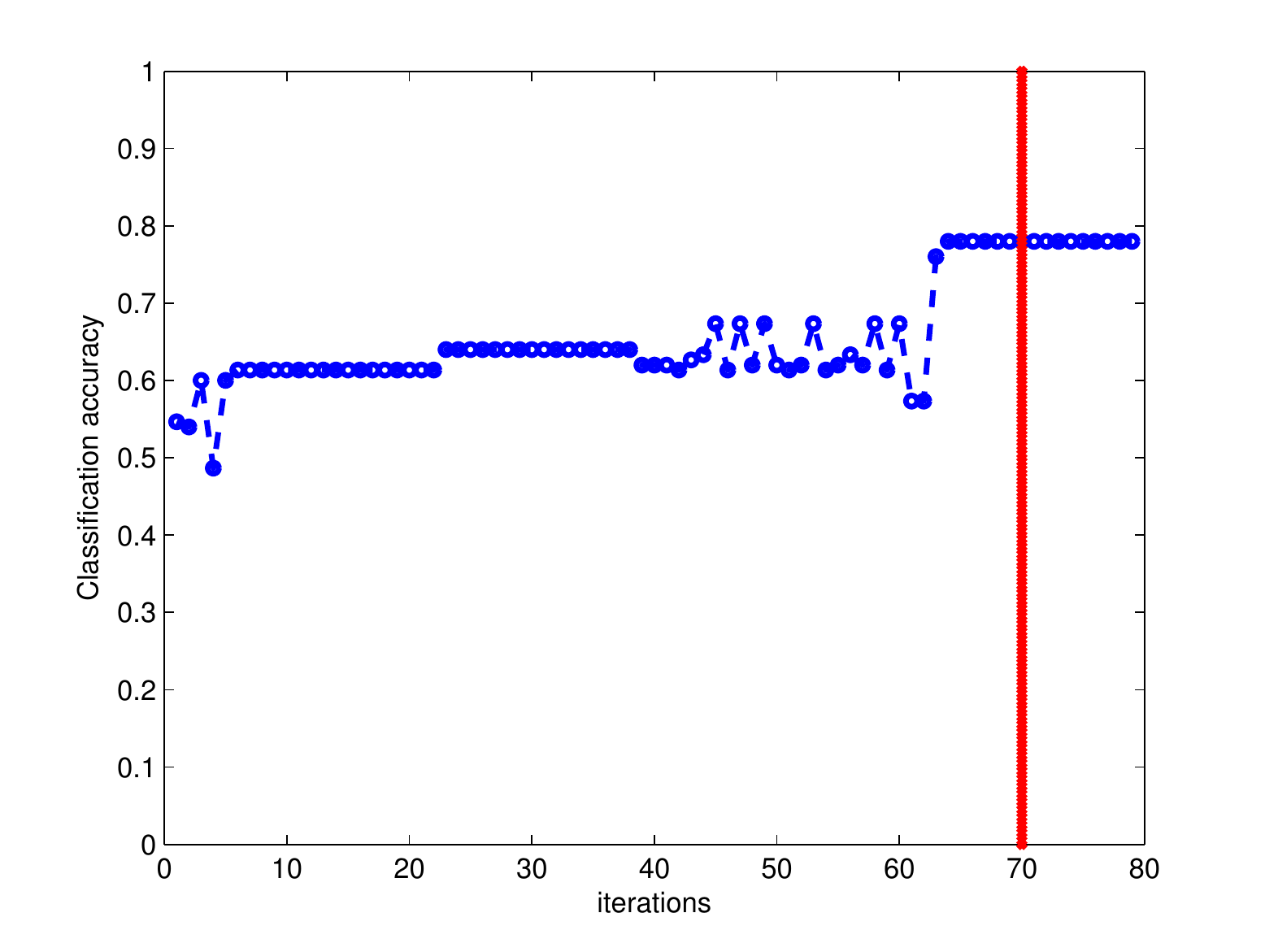}
  \caption{D $\rightarrow$ A}  
  \label{fig:4}
\end{subfigure}%
\begin{subfigure}[b]{0.16\textwidth}
  \centering
  \includegraphics[width=\linewidth]{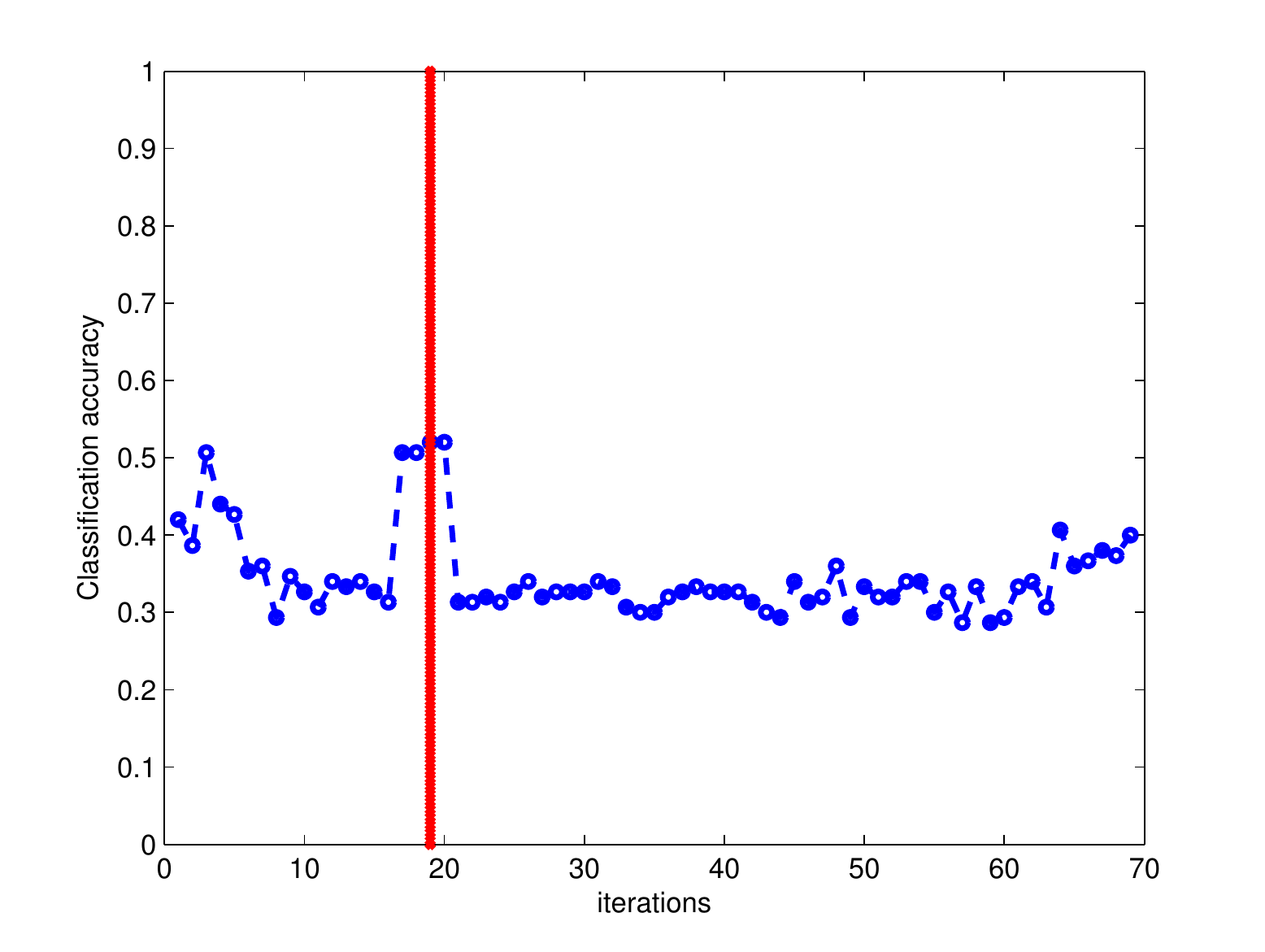} 
  \caption{D $\rightarrow$ C}
  \label{fig:5}
\end{subfigure}
\begin{subfigure}[b]{0.16\textwidth}
  \centering
  \includegraphics[width=\linewidth]{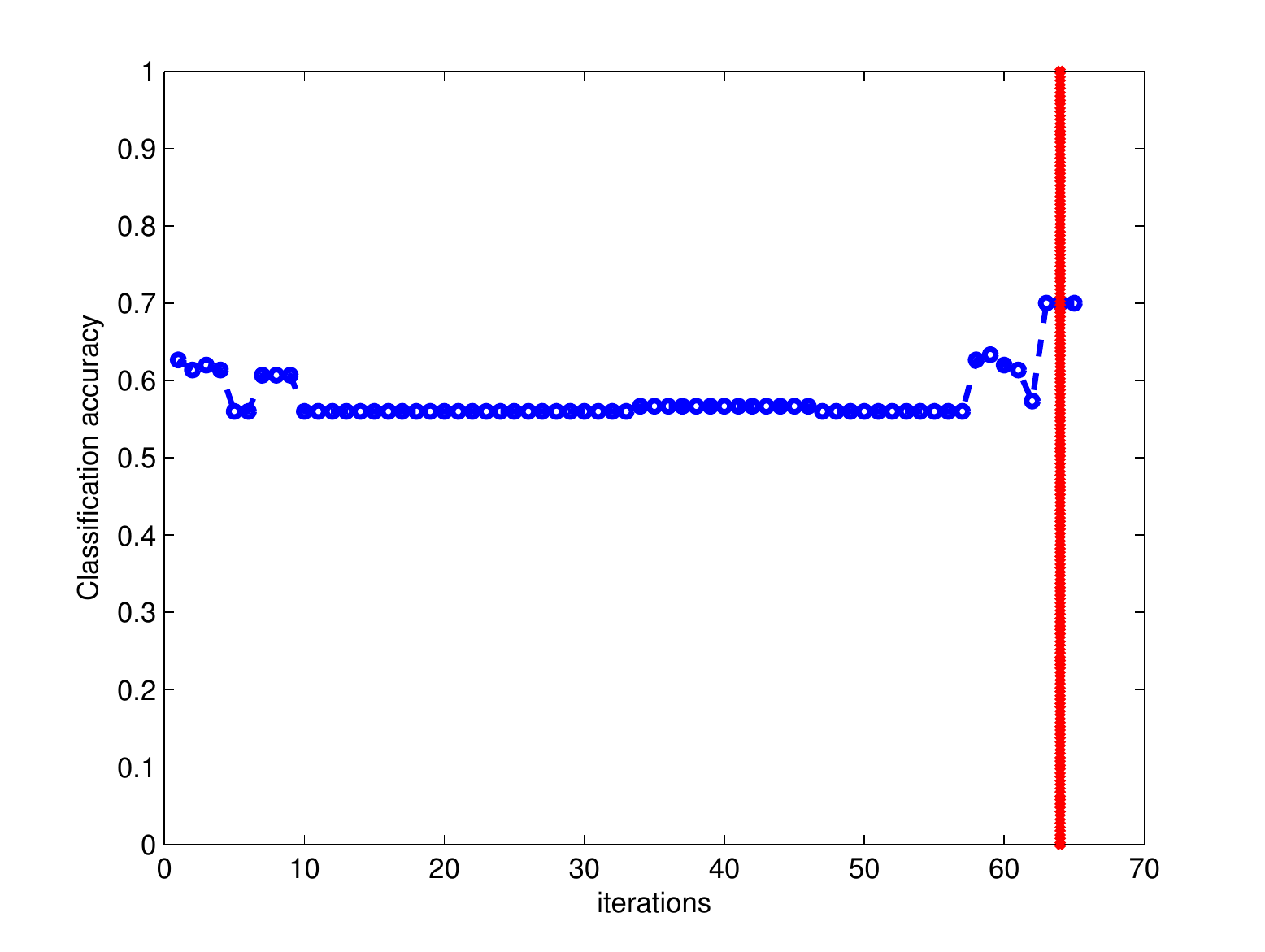}  
  \caption{D $\rightarrow$ W}
  \label{fig:6}
\end{subfigure}

\caption{Algorithm performance on 6 transfer learning scenarios.}
\vspace{-5mm}
\end{figure}

\section{Conclusions and future work}
In this paper we presented a new method for unsupervised transfer learning. We use kernel alignment optimization in order to minimize the distance between the distributions of source and target tasks. We apply K-NMF to the intermediate kernels obtained during this procedure and look for a weight matrix that reconstructs well the similarity based representation of data. Once this matrix is found, we use it in C-NMF on the target task to obtain the final partition. Our approach was evaluated on benchmark computer vision data sets and demonstrated a significant improvement when compared to some state-of-art methods. We also showed how KTA maximization can be related to HSIC and QMI optimization. The established relationships allow us to conclude that the use of KTA for transfer learning is justified from both theoretical and practical points of view. One of the inconvenients of our approach is that it is quite time consuming. Nevertheless, this issue can be overcome as discussed in section 3. 

In future, we will extend our work in the multiple directions. First of all, we will start by creating a multi-task version of our method. This can be done in the same fashion but with the only difference: firstly, we will search an optimal Gram matrix for each pair of tasks, then we will use the simultaneous non-negative matrix factorization \cite{badea} to find the common ``bridge matrix" that captures the knowledge from all tasks. Multi-task version of our algorithm can be very important because it could show us the participation of each task in overall improvement. Secondly, it would be useful to derive bounds for classification error. This problem, however, is complicated as there is no statistical theory that can be used in unsupervised setting in the same way how it can be done for supervised and semi-supervised learning.    




\begin{thebibliography}{9}

\bibitem{panyang}
  Pan, S.J. and Yang, Q.
  \emph{A Survey on Transfer Learning}.
  IEEE Transactions on Knowledge and Data Engineering, pp. 1345-1359, 2010. 

\bibitem{daiyang}
  Dai, Wenyuan and 0001, Qiang Yang and Xue, Gui-Rong and Yu, Yong.
  \emph{Self-taught clustering}.
  Proceedings of ICML, pp. 200-207, 2008. 

\bibitem{jiang} Wenhao Jiang and Fu-Lai Chung. 
        {\it Transfer Spectral Clustering}, Proceedings of the ECML/PKDD, pp. 789-803, 2012.

\bibitem{journals/ijon/ZhangZ11}
  Zhang, Jianwen and Zhang, Changshui.
  \emph{Multitask Bregman clustering}.
  Neurocomputing, pp. 1720-1734, 2011.

\bibitem{leeseung}
  D.D. Lee, H.S. Seung.
  \emph{Learning the parts of objects by non-negative matrix factorization}.
  Nature 401, 788 - 791, 1999. 
  
\bibitem{dingli}
  Ding, Chris H. Q. and Li, Tao and Jordan, Michael I.
  \emph{Convex and Semi-Nonnegative Matrix Factorizations}.
  IEEE Trans. Pattern Anal. Mach. Intell., vol. 32, pp. 45-55, 2010.  

\bibitem{zhangchen}
  Zhang, D., Zhou, Z.H. and Chen, S.
  \emph{Non-negative matrix factorization on kernels}.
  Proceedings of the 9th Pacific Rim International Conference on Artificial Intelligence, pp. 404-412, 2006.

\bibitem{crist}
  Cristianini, N., Shawe-Taylor, J. , Elisseeff, A. and Kandola, J.
  \emph{On kernel-target alignment}.
  NIPS, pp. 367-373, 2002.

\bibitem{rendon}
  Erendira Rendon, Itzel Abundez, Alejandra Arizmendi and Elvia M. Quiroz.
  \emph{Internal versus external cluster validation indexes}.
  International Journal of Computers and Communications, vol. 5, no. 1, 2011.

\bibitem{kr:2008RSA}
  Kriegeskorte, N. and Mur, M. and Bandettini, P.
  \emph{Representational similarity analysis-connecting the branches of systems neuroscience}.
  Frontiers in systems neuroscience, pp. 1-28, 2008.

\bibitem{Neumann-et-al-ML}
  Neumann, J. and Schn{\"o}rr, C. and Steidl, G.
  \emph{{C}ombined {SVM}-based {F}eature {S}election and {C}lassification}.
  Machine Learning, vol. 61, pp. 129-150, 2005.
  
\bibitem{ramona}
  Ramona, M., Richard, G. and David, B.
  \emph{Multiclass Feature Selection with Kernel Gram-matrix-based criteria}.
  IEEE Trans. Neural Netw. Learning Syst., 2012.

\bibitem{pothin}
  Pothin, J.-B., and Richard, C.
  \emph{A greedy algorithm for optimizing the kernel alignment and the performance of kernel machines}.
  In Proc. EUSIPCO ’06, pp. 4-8, 2006.

\bibitem{Gretton:2005:MSD:2101372.2101382}
  Gretton, Arthur and Bousquet, Olivier and Smola, Alex and Sch\"{o}lkopf, Bernhard.
  \emph{Measuring Statistical Dependence with Hilbert-schmidt Norms}.
  Proceedings of ALT, pp. 63--77, 2005.

\bibitem{icml2013_zhang13d}
  Kun Zhang and Bernhard Sch{\"o}lkopf and Krikamol Muandet and Zhikun Wang.
  \emph{Domain Adaptation under Target and Conditional Shift}.
  Proceedings of ICML, pp. 819-827, 2013.

\bibitem{principe}
  Principe, Jose C.
  \emph{Information Theoretic Learning: Renyi's Entropy and Kernel Perspectives}.
  Springer Publishing Company Incorporated, 2010.


\bibitem{parzen1962estimation}
  Parzen, Emanuel.
  \emph{On Estimation of a Probability Density Function and Mode}.
  The Annals of Mathematical Statistics, vol. 33, pp. 1065-1076, 1962.

\bibitem{GongGS13}
  Boqing Gong and Kristen Grauman and Fei Sha.
  \emph{Connecting the Dots with Landmarks: Discriminatively Learning Domain-Invariant Features for Unsupervised Domain Adaptation}.
  Proceedings of ICML, pp. 222-230, 2013.
  
    \bibitem{dingjor}
  Li, Tao and Ding, Chris H. Q. and Jordan, Michael I..
  \emph{Solving Consensus and Semi-supervised Clustering Problems Using Nonnegative Matrix Factorization}.
  Proceedings of ICDM, pp. 577-582, 2007.

  
\bibitem{Saenko:2010:AVC:1888089.1888106}
  Saenko, Kate and Kulis, Brian and Fritz, Mario and Darrell, Trevor.
  \emph{Adapting Visual Category Models to New Domains}.
  Proceedings of ICCV, pp. 213--226, 2010.

\bibitem{Gopalan:2011:DAO}
  Gopalan, Raghuraman and Ruonan Li and Chellappa, Rama.
  \emph{Domain Adaptation for Object Recognition: An Unsupervised Approach}.
  Proceedings of ICCV, pp. 999--1006, 2011.

\bibitem{badea}
  Badea, Liviu.
  \emph{Extracting Gene Expression Profiles Common to Colon and Pancreatic Adenocarcinoma Using Simultaneous Nonnegative Matrix Factorization}.
  World Scientific, pp. 267-278, 2008.

\end{thebibliography}
\end{document}